\documentclass[3p]{elsarticle}

\makeatletter
\def\ps@pprintTitle{%
 \let\@oddhead\@empty
 \let\@evenhead\@empty
 \def\@oddfoot{}%
 \let\@evenfoot\@oddfoot}
\makeatother

\usepackage[utf8]{inputenc}
\usepackage[T1]{fontenc}
\usepackage{graphicx}
\usepackage{float}                      
\usepackage{geometry}                   
\usepackage{setspace}                   
\usepackage{color}                      
\usepackage[table]{xcolor}              
\usepackage{fancyhdr}                   

\sffamily                               

\title{Unit contradiction versus unit propagation}
\author[]{Olivier Bailleux}
\ead{olivier.bailleux@u-bourgogne.fr}
\address{Université de Bourgogne, Faculté des Sciences et Techniques, département IEM\\ BP47870, 21078 Dijon \textsc{cedex}}

\date{}
\usepackage{amssymb}
\usepackage{amstext}
\usepackage[normalem]{ulem}

\newtheorem{thm}{Theorem}
\newproof{pf}{Proof}
\usepackage{amsmath}

\begin{document}

\begin{abstract}
Some aspects of the result of applying unit resolution on a \textsc{cnf} formula can be formalized as functions with domain a set of partial truth assignments. We are interested in two ways for computing such functions, depending on whether the result is the production of the empty clause or the assignment of a variable with a given truth value. 
We show that these two models  can compute the same  functions with formulae of polynomially related sizes, and we explain how this   result is related to the \textsc{cnf} encoding of Boolean constraints.
\end{abstract}

\sloppy 


\maketitle



\section{Introduction}

\subsection{Theoretical framework}

In this paper, we deal with Boolean variables, constraints, and assignments.
Any assignment of a Boolean variable $v$ is denoted either $[v,0]$ or $[v,1]$.  Given any set $V$ of Boolean variables, an \emph{assignment on $V$} is a set $I$ of assignments of variables of  $V$. $I$ is said to be \emph{complete} if it assigns exactly one value to any variable of $V$, \emph{partial} if it assigns at most one value to any variable of $V$, and \emph{contradictory} if there is a variable $v \in V$ such that $[v,0] \in V$ and $[v,1] \in V$. Unless otherwise stated, an assignment on $V$ is supposed to be partial and not contradictory. The set of all possible non contradictory complete assignments on $V$ will be denoted $\mathcal{A}_V$, while the set of all possible non contradictory partial assignments on $V$ will be denoted $\mathcal{I}_V$.

Given any set $V$ of Boolean variables, the term \emph{Boolean constraint on $V$} will be used in its widest sense, i.e.,  any computational representation $q$ of a \emph{satisfiability function} $h_q$ with domain $\mathcal{A}_V$ and codomain $\{ \mathtt{sat}, \mathtt{unsat} \}$. 
Given any complete assignment $A \in \mathcal{A}_V$, any constraint $q$ on $V$ is said to be \emph{satisfied} by $A$ if $h_q(A) = \mathtt{sat}$, else it is said to be \emph{falsified} by $A$. Given any partial assignment $I \in \mathcal{I}_V$, any constraint $q$ on $V$ is said to be satisfied (falsified, respectively) by $I$ if and only if $q$ is satisfied (falsified, respectively) by any $A \in \mathcal{A}_V$ such that $I \subseteq A$.

In propositional logic, a literal is either a propositional variable $v$ or its negation $\neg v$. By convention, the truth values will be denoted as the Boolean values 0 and 1. A clause is any disjunction of literals $\omega_1 \vee \cdots \vee \omega_k$, and a \textsc{cnf} formula is any conjunction of clauses $c_1 \wedge \cdots \wedge c_m$. The \emph{size of a clause} is its number of literals. The \emph{size of a formula} is the sum of the sizes of its clauses.

Literals, clauses, and \textsc{cnf} formulae can be considered as Boolean constraints: $v$ is satisfied by $[v,1]$, $\neg v$ by $[v,0]$, a clause is satisfied if and only if at least one of its literals is satisfied, and a \textsc{cnf} formula is satisfied if and only if all its clauses are satisfied. 

The following conventions will be used in the rest of the paper: given any variable $v$, the assignment $[v,1]$ can be denoted $[v]$, and the assignment $[v,0]$ can be denoted $[\neg v]$; any clause can be considered as a set of literals, and any formula can be considered as a set of clauses; for any set $V$ of Boolean variables, $\mathtt{ lit}(V)$ denotes the set of literals based on variables of $V$, namely $\cup_{v \in V}{\{ v, \neg v\}}$.

Any \textsc{cnf} formula $\Sigma$ is said to be \emph{satisfiable} if and only if there exists a truth assignment which satisfies $\Sigma$.
\textsc{Sat} is the problem of determining whether any arbitrary \textsc{cnf} formula $\Sigma$ is satisfiable or not.
Given any formula $\Sigma$ with variables $V$, and any  assignment $I$ on $V$, $\Sigma|_I$ denotes the formula $\Sigma \wedge_{[\omega] \in I}{(\omega)}$, i.e., the formula $\Sigma$ where the clause $(v)$ is added for each assignment of $[v,1] \in I$, and the clause $(\neg v)$ is added for each assignment $[v, 0] \in I$.

Introduced in \cite{unit-resol-64}, \emph{unit resolution} utilizes \emph{unit clauses} to produce new variable assignments and, when applicable, to detect inconsistencies. For the purpose of this paper, its principle can be described as follows. Given any assignment $I$, a clause $c$ is said to be a unit clause with respect to $I$ if and only if $I$ falsifies all the literals of $c$ except for one literal $\omega$, which will be called the \emph{active literal} of $c$. Given any \textsc{cnf} formula $\Sigma$, the unit resolution process starts from an empty set $U$ of variable assignments, which is iteratively augmented by the active literals of unit clauses with respect to $U$, until either $U$ becomes contradictory or no new literal can be inferred any more. The formula can then be simplified by removing any non-unit clause satisfied by $U$, as well as any literal falsified by $U$. The resulting formula $\Sigma'$ is logically equivalent to $\Sigma$. If $U$ is contradictory, then the empty clause belongs to $\Sigma'$, implying that $\Sigma$ is not satisfiable.
\textsc{Sat} solvers \cite{sat-handbook} use unit resolution to speed up the search for solutions or inconsistencies by reducing the number of decisions (binary nodes) in the search tree.

\subsection{Motivation}

Given any set $V$ of propositional variables, we are interesting in functions with domain $D \subseteq \mathcal{I}_V $ and codomain $\{\mathtt{yes}, \mathtt{no} \}$\footnote{Without loss of generality, these values have been chosen so as to avoid any ambiguity with the logical values \texttt{true} and \texttt{false} or the Boolean values 0 and 1.} which can account for some aspects of the result of applying unit resolution to a \textsc{cnf} formula: the empty clause is produced, or a given variable is assigned to 1, or it is assigned to 0. In the scope of this report, these functions will be called \emph{matching functions}.

Given any formula $\Sigma$, and any set $V$ of propositional variables occurring in $\Sigma$, the inferences made by unit resolution can be modeled by the following matching functions:
\begin{itemize}
\item
The function $f_{\Sigma} : \mathcal{I}_V \mapsto \{\mathtt{yes},\mathtt{no}\}$ such that for any partial assignment $I \in \mathcal{I}_V$, $f_{\Sigma}(I)=\mathtt{yes}$ if and only if applying unit resolution on $\Sigma|_I$ produces the empty clause. We will say that unit resolution computes this function \emph{by contradiction}.
\item
For any literal $\omega = \mathtt{lit}(V)$, the function $g_{\Sigma, \omega} : D_{\omega} \mapsto \{\mathtt{yes}, \mathtt{no}\}$, where $D_{\omega} = \{I \in \mathcal{I}_V\ : f_{\Sigma}(I) = \mathtt{no}\}$, such that for any partial truth assignment $I \in D_{\omega} $, $g_{\Sigma, \omega}(I) = \mathtt{yes}$ if and only if applying unit resolution to $\Sigma|_I$ infers $[\omega]$. We will say that unit resolution computes these functions \emph{by propagation}.
\end{itemize}

Knowing the matching functions that can be computed by contradiction -- as well as the ones that can be computed by propagation --  with a polynomial amount of clauses  is crucial for the study of the \textsc{cnf} encodings of Boolean constraints. Given any set $V$ of Boolean variables and any constraint $q$ on $V$, a \textsc{cnf} encoding of $q$ is any \textsc{cnf} formula $\Sigma_q$ (which can include variables not belonging to $V$) such that for any \emph{complete} assignment $A \in \mathcal{A}_V$, $\Sigma_q|_A$ is satisfiable if and only if $A$ satisfies $q$. This property  allows
any constraint satisfiability problem to be solved using a \textsc{sat} solver.

Two interesting additional properties of \textsc{cnf} encodings have been reported as potentially improving the efficiency of solving the resulting \textsc{sat} instances:

\begin{enumerate}
\item
Given any encoded constraint $q$, unit resolution detects any \emph{partial} assignment which falsify $q$: from any such assignment, the empty clause is produced. For example, this property is studied in \cite{sinz-05} in the context of Boolean cardinality constraints. Such an encoding will be called \texttt{upi} (like unit propagation detects inconsistency) in the following.
\item
Given any encoded constraint $q$ on variables $V$, unit resolution enforces the generalised arc consistency of $q$, i.e., for any partial assignment $I \in \mathcal{I}_V$ which does not falsify $q$, and any literal $\omega \in \mathtt{lit}(V)$, if $I \cup \{ [\omega] \}$ falsify $q$ then $[ \neg \omega]$ is inferred. This criterion was introduced in  \cite{gent-02}. Such an encoding will be called \texttt{upac} (like unit propagation restores generalized arc consistency) in the following.
\end{enumerate}

In most cases, only the encodings producing a formula of size polynomially related to the number of variables of the input constraint can be used in practice. They will be called \emph{polynomial encodings} in the following.

Let us consider a family $\mathcal{Q}$ of constraints on Boolean variables. For any constraint $q$ of $\mathcal{Q}$ with variables $V = \{ v_1, \ldots, v_n \}$, let us define the \emph{inconsistency function} of $q$ as $f_q : \mathcal{I}_V \mapsto \{\mathtt{yes},\mathtt{no}\}$ such that for any \emph{partial} assignment $I \in \mathcal{I}_V$, $f_q(I)=\mathtt{yes}$ if and only $I$ falsifies $q$.

Clearly, the existence of a polynomial \texttt{upi} encoding for the constraints of $\mathcal{Q}$ depends on the existence of polynomially sized \textsc{cnf} formulae allowing unit resolution to compute \emph{by contradiction} the inconsistency functions of the constraints of $\mathcal{Q}$.

Now, for any constraint $q$ of $\mathcal{Q}$ with variables $V = \{ v_1, \ldots, v_n \}$, and any literal $\omega \in \mathtt{lit}(V)$, let us define the \emph{arc consistency functions} of $q$ as $g_{q,\omega} : D_q  \mapsto \{\mathtt{yes},\mathtt{no}\}$, $D_q = \{I \in \mathcal{I}_V : f_q(I) = \mathtt{no} \}$, such that for any partial assignment $I \in D_q$, $g_{q,\omega}(I )= \mathtt{yes}$ if and only if $I \cup \{ [\neg \omega] \}$ falsifies $q$.


Clearly again, the existence of a polynomial \texttt{upac} encoding for the constraints of $\mathcal{Q}$ can be expressed as the existence of polynomially sized \textsc{cnf} formulae allowing unit resolution to compute \emph{by propagation} the arc consistency functions related to the constraints of $\mathcal{Q}$.

\subsection{Contribution}

We show that any matching function can be computed by unit contradiction if and only if it can be computed by unit propagation, and that any family of matching functions can be computed in polynomial size (then in polynomial time) by unit contradiction if and only if it can be computed in polynomial size by unit propagation.

As a corollary, for any family $\mathcal{Q}$ of constraints with Boolean variables, if there exists a polynomial \texttt{upi} encoding for $\mathcal{Q}$ then there exists a polynomial \texttt{upac} encoding for $\mathcal{Q}$.

\section{Technical results\label{theorems}}

In this section, we will formalize and prove the previously presented results as the two following theorems.

\begin{thm}{}\label{up2uc}
Let $f$ be any matching function.
If $f$ can be computed by propagation using a formula of size $p$, then $f$ can be computed by contradiction with a formula of size $p+1$.
\end{thm}

\begin{pf}
Any \textsc{cnf} formula $\Sigma_p$ computing a matching function $f$ by propagation can be reduced in the following way to a formula $\Sigma_c$ computing $f$ by contradiction.

Let $f$ be a matching function with domain $D \subseteq \mathcal{I}_V$. Let $\Sigma_p$ be a \textsc{cnf} formula allowing unit resolution to compute $f$ by propagation. This means that there is a literal $\omega$ such that for any $I \in D$, applying unit propagation to $\Sigma_p|_I$ cannot produce the empty clause, but infers $[\omega]$ if and only if $f(I)=\mathtt{yes}$. Then the formula $\Sigma_c = \Sigma_p \wedge (\neg \omega)$ allows unit resolution to compute $f$ by contradiction.
$\square$
\end{pf}

\begin{thm}{}\label{uc2up}
Let $f$ be any matching function. If $f$ can be computed by contradiction using a formula of size $p$ with $n$ variables, then $f$ can be computed by propagation with a formula of size $O(p n^2)$.
\end{thm}

\begin{figure}[h]
\begin{center}
\includegraphics[scale=0.7]{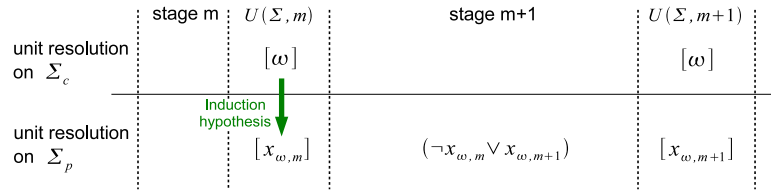}
\caption{The case 1 of the proof of the induction hypothesis $H_m$\label{proof-part1}}
\end{center}
\end{figure}

\begin{figure}[h]
\begin{center}
\includegraphics[scale=0.7]{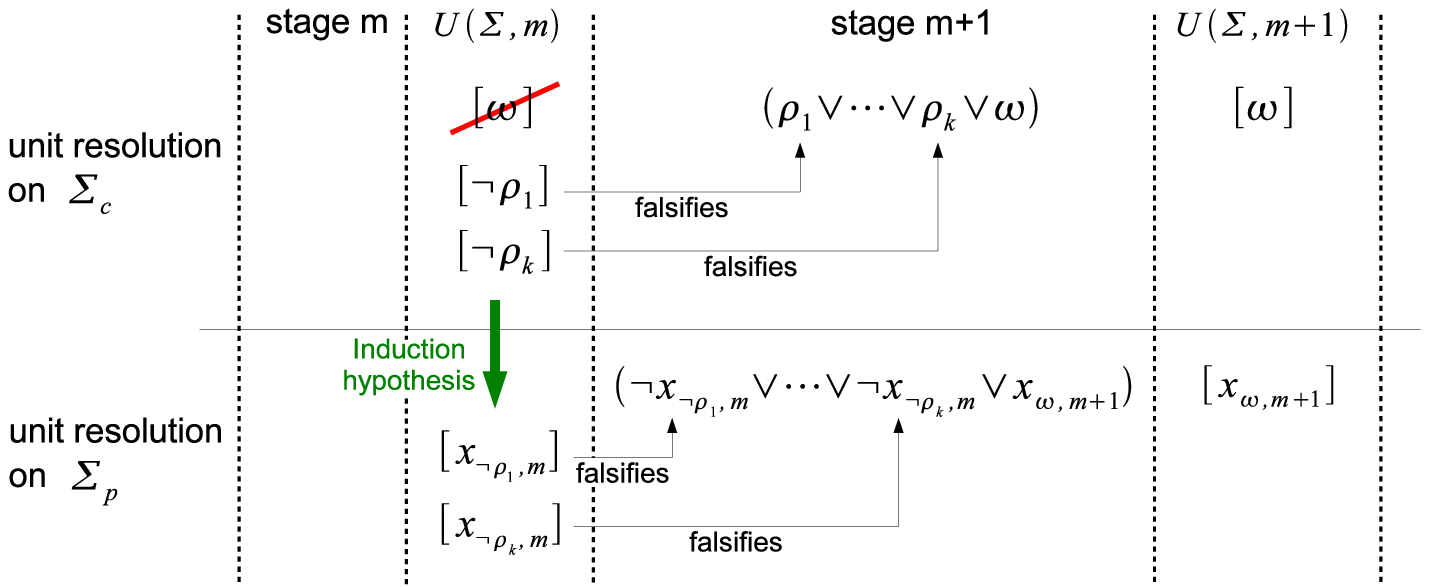}
\caption{The case 2 of the proof of the induction hypothesis $H_m$\label{proof-part2}}
\end{center}
\end{figure}

\begin{pf}
Any \textsc{cnf} formula $\Sigma_c$ computing a matching function $f$ by contradiction can be reduced in the following way to a formula $\Sigma_p$ computing $f$ by propagation.

Let $f$ be a matching function with domain $\mathcal{I}_V$. Let $\Sigma_c$ be a \textsc{cnf} formula allowing unit resolution to compute $f$ by contradiction. We will construct a formula $\Sigma_p$ such that for any $I \in \mathcal{I}_V$, applying unit resolution to $\Sigma_p|_I$ does not produce the empty clause, but assigns 1 to a new variable $s$ if and only if applying unit resolution to $\Sigma_c|_I$ produces the empty clause. As a manner of speaking, applying unit resolution on $\Sigma_p|_I$ simulates the effects of applying unit resolution to $\Sigma_c|_I$ without ever producing the empty clause.

To this end, the unit resolution process is decomposed into stages such that each stage $i$ produces the assignments induced from the unit clauses with respect to the   assignment of the stage $i-1$, where the  assignment of the stage $0$ is $I_0 = I$. Let $V$ be the set of variables of $\Sigma_c$ and $n = |V|$. Because the cardinal of any non contradictory  assignment on $V$ is at most $n$, the unit resolution process stops after at most $n+1$ stages.

Given any \textsc{cnf} formula $\Sigma$ and any integer $i$, let $\mathtt{U}(\Sigma,i)$ denote the current  assignment after $i$ unit resolution stages on $\Sigma$.

The formula $\Sigma_p$ contains $2(n+1)+n$ variables, namely the variables of $V$ and, for each literal $\omega \in \mathtt{lit}(V)$, $(n+1)$ new variables denoted $x_{\omega,1}, \ldots, x_{\omega,n+1}$. It consists of the following clauses:
\begin{enumerate}
\item
for any $v \in V$, $(v \vee x_{\neg v,1})$ and $(\neg v \vee x_{v,1})$, which are called \emph{injection clauses};
\item
for any $\omega \in \mathtt{lit}(V)$ and any $i \in 1..n$, $(\neg x_{\omega,i} \vee x_{\omega, i+1})$, which are called  \emph{replication clauses};
\item
for any clause $c$ of $\Sigma_c$ with at least two literals, any literal $\omega  \in c$, and any $i \in 1..n,$ $(x_{\omega, i+1} \vee_{\rho \in c \setminus \{ \omega \}}{\neg x_{\neg \rho, i}})$, which are called \emph{deduction clauses}.
\item
for any singleton clause $(\omega)$ of $\Sigma$, $(x_{\omega, 1})$, which are called \emph{unit clauses}.
\end{enumerate}

Let us consider the following induction hypothesis $H_m$: for any $\omega \in \mathtt{lit}(V)$, $[x_{\omega, m}] \in \mathtt{U}(\Sigma_p|_I,m)$ if and only if $[\omega] \in \mathtt{U}(\Sigma_c|_I,m)$.

For any $\omega \in \mathtt{lit}(V)$, $[\omega] \in \mathtt{U}(\Sigma_c|_I,1)$  if and only if $[\omega] \in I$ or $(\omega) \in \Sigma_c$. In the first case, $[x_{\omega, 1}] \in \mathtt{U}(\Sigma_p|_I,1)$  thanks to the injection clause $(\neg \omega \vee x_{\omega, 1})$. In the second case, $[x_{\omega, 1}] \in \mathtt{U}(\Sigma_p|_I,1)$ thanks to the  deduction clause $(x_{\omega, 1})$. Because only these clauses can infer $[x_{\omega, 1}]$  during the unit resolution process on $\Sigma_p|_I$, and because they can infer $[x_{\omega, 1}]$ only in theses two cases, $H_1$ holds.

Now, suppose that $H_m$ holds for some $m \in 1..n$, and let us consider any literal $\omega \in \mathtt{lit}(V)$. Regarding the inference of $[\omega]$  by unit resolution on $\Sigma_c|_I$ at stage $m+1$, three cases can be considered:
\begin{enumerate}
\item
$[\omega] \in \mathtt{U}(\Sigma_c|_I,m)$ and then $[\omega] \in \mathtt{U}(\Sigma_c|_I,m+1)$. By induction hypothesis, $[x_{\omega, m}] \in \mathtt{U}(\Sigma_p|_I,m)$. Thanks to the replication clause $(\neg x_{ \omega, m}, \vee x_{ \omega, m+1})$ of $\Sigma_p$, $[ x_{ \omega, m+1}] \in \mathtt{U}(\Sigma_p|_I,m+1)$. See Figure \ref{proof-part1} for a graphical illustration.

\item
$[\omega] \notin \mathtt{U}(\Sigma_c|_I,m)$ and $[\omega] \in \mathtt{U}(\Sigma_c|_I,m+1)$. Then there is a clause $(\rho_1, \vee \cdots \vee \rho_k \vee \omega)$ in $\Sigma_c$ such that all the assignments $[\neg \rho_1]$ to $[\neg \rho_k]$ are in $\mathtt{U}(\Sigma_c|_I,m)$. By induction hypothesis, $[x_{\neg \rho_1, m}]$ to $[x_{\neg \rho_k, m}]$ are in $\mathtt{U}(\Sigma_p|_I,m)$. Thanks to the deduction clause $(\neg x_{\neg \rho_1, m} \vee \cdots \vee \neg x_{\neg \rho_k, m} \vee x_{\omega, m+1})$, $[x_{\omega, m+1}] \in \mathtt{U}(\Sigma_p|_I,m+1)$. See Figure \ref{proof-part2} for a graphical illustration.
\item
$[\omega] \notin \mathtt{U}(\Sigma_c|_I,m)$ and $[\omega] \notin \mathtt{U}(\Sigma_c|_I,m+1)$. The only clauses of $\Sigma_p$ that can infer $[x_{\omega, m+1}]$ are the replication clauses and the deduction clauses. By induction hypothesis, $[x_{\omega, m}] \notin \mathtt{U}(\Sigma_p|_I,m)$, then no replication clause can infer $[x_{\omega, m+1}]$. Secondly, because $[\omega] \notin \mathtt{U}(\Sigma_c|_I,m+1)$, for any clause $(\rho_1, \vee \cdots \vee \rho_k \vee \omega)$ in $\Sigma_c$, not all the literals $\rho_1$ to $\rho_k$ are falsified by $U(\Sigma_c|_I,m)$. By induction hypothesis, not all the assignments $[x_{\neg \rho_1, m}]$ to $[x_{\neg \rho_k, m}]$ are in $\mathtt{U}(\Sigma_p|_I,m)$. Then, the corresponding deduction clause $(\neg x_{\neg \rho_1, m} \vee \cdots \vee \neg x_{\neg \rho_k, m} \vee x_{\omega, m+1})$ of $\Sigma_p$ cannot infer $[x_{\omega, m+1}]$.
\end{enumerate}

Hence $H_m$ holds for any $m \in 1..(n+1)$. Furthermore, because each unit resolution stage on $\Sigma_p|_I$ infers only positive literals, the empty clause is never produced. It follows that unit resolution on $\Sigma_c|I$ produces the empty clause (or, equivalently, infers two opposite literals) if and only if there is a variable $v \in V$ such that $[x_{v, n+1}]$ and $[x_{\neg v, n+1}]$ are inferred by unit resolution on $\Sigma_p|_I$. Now, let us complete the formula $\Sigma_p$ with the clauses $(\neg x_{v,n+1} \vee \neg v_{\neg v, n+1} \vee s)$, for each $v \in V$, where $s$ is a new fresh variable. Clearly, unit resolution on $\Sigma_p|_I$ infers $[s]$ if and only if unit resolution on $\Sigma_c|_I$ produces the empty clause. For illustrative purposes, Figure \ref{exp-reduction} gives an example of how unit resolution on $\Sigma_p|_I$ simulates unit resolution on $\Sigma_c|_I$.

Let $p$ be the size of $\Sigma_c$ and $k$ be the size of the largest clause of $\Sigma_c$. Without loss of generality, let us suppose that $k \leq n$ (any larger clause would be a tautology). The formula $\Sigma_p$ includes $O(n)$ injection clauses, $O(n^2)$ replication clauses, $O(n^2)$ unit clauses, and $O(np)$ deduction clauses. Because the largest clauses of $\Sigma_p$, which are the deduction clauses, have size at most $k$, the size of $\Sigma_p$ is $O(npk) = O(n^2p)$.
$\square$
\end{pf}

As a corollary of theorem \ref{uc2up}, let $\mathcal{Q}$ be a family of contraints for which there exists a polynomial \texttt{upi} \textsc{cnf} encoding, and let us show how a polynomial \texttt{upac} encoding can be obtained.

By hypothesis, for any constraint $q \in \mathcal{Q}$ with variables $V = \{v_1, \ldots, v_n\}$, there is a \textsc{cnf} formula $\Sigma_q$ of size polynomially related to $n$ such that for any assignment $I$ on $V$, unit resolution on $\Sigma_q|_I$ produces the empty clause if and only if $I$ falsifies $q$.
A \texttt{upac} encoding $\Omega_q$ for $q$ must verify the additional following property: for any literal $\omega \in \mathtt{lit}(V)$, and any assignment $I \in \mathcal{ I}_V$ such that $I$ does not falsify $q$ and $[\omega] \notin I$, unit resolution on $\Omega_q|_I$ does not produce the empty clause, but infers $[\omega]$ if and only if $I \cup \{ [\neg \omega] \}$ falsifies $q$. Such a behavior can be obtained thanks to the following formula:

\[
\Omega_q = \Sigma_q \wedge_{\omega \in \mathtt{lit}(V)}{\left( \Sigma_{q, \omega} \wedge (\neg s_{\omega} \vee \neg \omega) \right)}
\]

Where each $\Sigma_{q, \omega}$ is a formula allowing unit resolution to compute \emph{by propagation}, with output variable $s_{\omega}$, the contradiction function of the contraint $\Sigma_q \wedge (\omega)$.

\section{Related works}

There are at least three research directions related to the study of the expressive power of unit resolution.

The first one aims to identify the classes of formulae for which unit resolution is a complete refutation procedure in the sense that it produces the empty clause if and only if the input formula is not satisfiable. For example, this property holds for the formulae containing only Horn clauses
\cite{henschen-74}. 

The second direction aims to characterize the complexity of determining whether a given formula can be refuted by unit resolution or not.
This decision problem denoted \textsc{unit} is known to be \textsc{p}-complete, meaning that for any decision problem $\pi$ with polynomial time complexity, there exists a $\log$ space reduction from $\pi$ to \textsc{unit} \cite{Jones1976}. Circuit value, which consists to determine the output value of a Boolean circuit, given its input values, is \textsc{p}-complete too \cite{goldschlager77}. Regarding the complexity theory, \textsc{unit} and circuit value have then the same expressive power.
In the present paper, a different point of view is adopted. The \textsc{cnf} formula is not the input data of a program, but the program itself. The input data is a \emph{partial} truth assignment encoded in a natural way, i.e., each input variable can be either assigned to 0, assigned to 1, or not assigned.

The third line is related to the search for efficient \textsc{cnf} encodings of various problems in order to solve them thanks to any \textsc{sat} solver. Because unit resolution is implemented efficiently in \textsc{sat} solvers, many works aim to find encoding schemes which allow unit resolution to make as many inferences as possible. In \cite{gent-02}, a \textsc{cnf} encoding for enumerative constraints is proposed, which allows unit propagation to make the same deductions on the resulting formula as restoring arc consistency on the initial constraints does. This work was innovative because with the previously known encodings, unit propagation had less inference power than restoring arc consistency, which is the basic filtering method used in constraint solvers \cite{csp-lecoutre}. It has been followed by various similar works on other kinds of constraints such as Boolean cardinality constraints \cite{bailleux-boufkhad-2003} and pseudo-Boolean constraints \cite{bailleux-boufkhad-roussel-2009}, for which polynomial \texttt{upi} and \texttt{upac} encoding are proposed. In \cite{Bacchus07a}, a general way to construct a (possibly non-polynomial) \texttt{upac} encoding for any constraint is proposed. Today, it has become customary, when a new encoding is proposed, to address the question of the behavior of unit resolution on the obtained \textsc{sat} instances. 
So far, \texttt{upi} and \texttt{upac} \textsc{cnf} encodings are known for enumerative constraints \cite{gent-02}, Boolean cardinality constraints \cite{sinz-05}, and pseudo-Boolean constraints \cite{gacpseudoboo}, but the research field remains open regarding, for example, arithmetic constraints \cite{booleanization} or global cardinality constraints \cite{regin-96}. 

\section{Concluding remarks and perspectives}

To the best of our knowledge, it is the first time that unit resolution is addressed as a computation model for functions with domain a set of \emph{partial} assignments on Boolean variables. We believe that this model is appropriate to characterize the inference power of unit resolution in \textsc{sat} solvers.
By showing that unit contradiction has the same expressive power as unit propagation, we provide a theoretical insight into the field of encodings of constraint satisfaction problems into \textsc{cnf} for solving them thanks to \textsc{sat} solvers. The underlying scientific issue is nothing less than determining the scope of application of \textsc{sat} solvers: which problems can be reasonably addressed by \textsc{sat} solvers, which cannot, and why ?

We are currently working on the characterization of the matching functions that can be efficiently computed by unit resolution, and so the constraints for which there exist polynomial \texttt{upi} and \texttt{upac} encodings.
The following step will be to look for a general method for translating -- when applicable -- algorithms or Boolean circuits into \textsc{cnf} formulae allowing unit resolution to compute the same matching functions.

\appendix

\section{Graphical illustrations}

Here, we give some graphical illustration of the reduction described in the proof of the theorems presented section \ref{theorems}.

\begin{figure}[H]
\begin{center}
\footnotesize
\begin{tabular}{|c|c|c|}\hline
 & unit resolution on $\Sigma_c|_I$ & unit resolution on $\Sigma_p|_I$\\
\hline
initial assignment $I$ & $[\neg b], [d]$ &  $[\neg b], [d]$\\
\hline
clauses invoked& $(a)$ & deduction clause: $(x_{a,1})$\\
at stage 1 & $(\neg b), (d)$ & injections clauses: $(b \vee x_{ \neg b,1}), (\neg d \vee x_{d,1})$\\
\hline
inferred assignments & $[a]$\ & $[x_{a,1}] , [x_{\neg b,1}], [x_{d,1}]$ \\
\hline
clauses invoked& $(\neg a \vee b \vee c)$ & deduction clause: $(\neg x_{a,1} \vee \neg x_{\neg b,1} \vee x_{c,2})$ \\
at stage 2 & $(\neg c \vee \neg d)$ & deduction clause: $(\neg x_{d,1} \vee x_{\neg c,2})$\\
& & replication clauses: $(\neg x_{a,1} \vee x_{a,2}), (\neg x_{\neg b,1} \vee x_{\neg b,2}), (\neg x_{d,1} \vee x_{d,2})$\\
\hline
inferred assignments & $[\neg c], [c]$ & $[x_{a,2}],[x_{\neg b,2}],[x_{d,2}],[x_{c,2}],[x_{\neg c,2}]$ \\
\hline
& unit resolution stops & unit resolution continues \\
 & due to contradiction & the inferred assignments $[x_{c,2}]$ and $[x_{\neg c,2}]$  \\
 &  & point out the contradiction in $\Sigma_c|_I$ \\
\hline
$\cdots$ & $\cdots$ & $\cdots$ \\
\hline
assignments after & & $[x_{a,5}],[x_{\neg b,5}],[x_{d,5}],[x_{c,5}],[x_{\neg c,5}]$ \\
stage 5 &  & (propagated thanks to the replication clauses) \\
\hline
clause invoked &  & $(\neg x_{c,5} \vee \neg x_{\neg c,5} \vee s)$ \\
at stage 6 &  &  \\
\hline
inferred assignment &  & $[s]$ \\
 &  & reifies the contradiction detected by unit resolution on $\Sigma_p|_I$ \\
\hline
\end{tabular}
\caption{Unit resolution on $\Sigma_c = (a) \wedge (\neg a \vee b \vee c) \wedge (\neg c \vee \neg d)$ and the corresponding formula $\Sigma_p$. \label{exp-reduction}}
\end{center}
\end{figure}

\bibliographystyle{plain}
\bibliography{biblio-propagation}

\end{document}